\documentclass[10pt,letterpaper,twocolumn]{article}
\pdfmapfile{+lm.map}
\pdfmapfile{+cm.map}
\pdfmapfile{+cmextra.map}
\pdfmapfile{+symbols.map}

\usepackage[T1]{fontenc}
\usepackage{lmodern}
\usepackage[letterpaper,margin=0.75in,columnsep=0.25in]{geometry}
\usepackage[hyphens]{url}
\usepackage{graphicx}
\usepackage{natbib}
\usepackage{booktabs}
\usepackage{amsmath,amssymb}
\usepackage{array}
\usepackage{multirow}
\usepackage{placeins}
\usepackage[hidelinks,hypertexnames=false]{hyperref}

\makeatletter
\g@addto@macro\@maketitle{\vskip 0.27em\relax}
\makeatother

\title{Do We Really Need Adaptive Global Spatial Attention for Traffic Forecasting?}
\author{%
Qihang Zhang\textsuperscript{1,2,3},
Siyao Zhang\textsuperscript{1,2,3},
Letao Kang\textsuperscript{1,2,3},\\
Wenzhe Liang\textsuperscript{1,2,3},
Miao Zhang\textsuperscript{4,5},
Zhao Zhang\textsuperscript{1,2,3,\(\dagger\)}\\[0.45em]
\normalsize \textsuperscript{1}School of Transportation Science and Engineering, Beihang University, Beijing, China\\
\normalsize \textsuperscript{2}Hangzhou International Innovation Institute, Beihang University, Hangzhou, China\\
\normalsize \textsuperscript{3}Key Laboratory of Intelligent Transportation Technology and System (Ministry of Education), Beijing, China\\
\normalsize \textsuperscript{4}School of Computer Science and Technology, Harbin Institute of Technology, Shenzhen, China\\
\normalsize \textsuperscript{5}Peng Cheng Laboratory, Shenzhen, China\\
\normalsize \textsuperscript{\(\dagger\)}Corresponding author: \href{mailto:zhaozhang@buaa.edu.cn}{zhaozhang@buaa.edu.cn}
}
\date{}

\hypersetup{
  pdftitle={Do We Really Need Adaptive Global Spatial Attention for Traffic Forecasting?},
  pdfauthor={Qihang Zhang, Siyao Zhang, Letao Kang, Wenzhe Liang, Miao Zhang, Zhao Zhang}
}

\begin{document}

\maketitle
\thispagestyle{empty}

\begin{abstract}
Existing traffic forecasting models commonly focus on extracting spatial dependencies, particularly global spatial information, which characterizes the representations obtained through interactions between each node and all nodes across the traffic network. However, the underlying mechanism by which global information is modeled and extracted remains insufficiently investigated. Whether global information must be extracted by high-degree-of-freedom adaptive attention or can be captured by a simple global aggregation operator remains unclear. For this purpose, we design a controlled ablation framework that replaces only the spatial mixing module to test attention-based global interaction. Across six traffic benchmarks, standard spatial attention yields relative MAE changes of $-1.58\%$ to $+1.26\%$ compared with uniform full-range mixing, and we observe no consistent advantage for standard spatial attention, while uniform full-range mixing reduces node-scale spatial-mixing complexity from $O(N^2)$ to $O(N)$. We further propose a hypothesized model that decomposes spatial attention into a row-uniform global background and a non-uniform residual. The residual shows dataset-dependent effects. Overall, uniform full-range mixing provides a strong global spatial baseline, while the non-uniform attention residual is not consistently beneficial across datasets.

\noindent\textbf{Code:} \url{https://github.com/uuesti/U-Trans}
\end{abstract}

\section{Introduction}

Traffic forecasting predicts future traffic states from historical observations collected by road sensors or spatial grids. In addition to short-term inertia and daily or weekly periodicity, traffic systems exhibit local propagation, regional co-movement, and network-wide demand or congestion patterns. In this work, global spatial information specifically refers to the representations acquired by each individual node through interactions with all nodes in the traffic network, thereby capturing network-level dependencies and shared traffic contexts beyond local neighborhoods. Recent traffic forecasting models commonly extract such broad spatial information through adaptive graphs, fully connected relations, spatial attention, dynamic long-range attention, or context broadcasting \citep{NEURIPS2020_ce1aad92,10.1145/3690624.3709177,Jiang_Han_Zhao_Wang_2023,lin2025multiview,Oreshkin_Amini_Coyle_Coates_2021,Zheng_Fan_Wang_Qi_2020}.

However, the use of global spatial information does not specify the required extractor. The key issue is whether the gain comes from a shared network-level background or from adaptive pairwise node-to-node weights. We focus on the attention-based aggregation mechanism and hold the remaining spatial-block components fixed through controlled variants.

Existing comparisons do not isolate this question. Graph propagation, adaptive adjacency, graph learning, and spatial attention have all improved traffic forecasting \citep{NEURIPS2020_ce1aad92,Jiang_Han_Zhao_Wang_2023,li2018diffusion,ijcai2019p264,10.1145/3394486.3403118,ijcai2018p0505,Zheng_Fan_Wang_Qi_2020}, but these models often change representation, temporal encoding, prediction head, and training protocol together. Strong representation-centric and simple baselines further highlight the importance of non-spatial components for forecasting performance \citep{10.1145/3583780.3615160,liu2023reallyneedgraphneural,10.1145/3511808.3557702,wang2026channel,10.1145/3583780.3614969}.

We use a representation-rich forecasting backbone and vary only the spatial aggregation module to study how the resulting multi-source node representations are spatially aggregated.

A closer but still under-examined case is global mean or low-degree-of-freedom global mixing. Recent traffic forecasting Transformers have introduced spatio-temporal context broadcasting to redistribute attention scores, making mean-style global background injection relevant to traffic forecasting \citep{lin2025multiview}. Yet such components are embedded in larger architectures and are not evaluated as isolated spatial mixers. Outside traffic forecasting, query-independent global context, uniform attention, Fourier token mixing, and other simple mixers suggest that low-degree-of-freedom global information transfer can be surprisingly effective \citep{9022134,Hyeon-Woo_2023_ICCV,lee-thorp-etal-2022-fnet}. Thus, global mean broadcasting should be treated as a serious baseline rather than a trivial ablation.

To address this identification problem, we use prediction ablations to test whether global mixing lowers forecasting error, and propose a hypothesized model that decomposes attention into the row mean and residual to explain the experimental findings.

The contributions are:

(1) We build a controlled framework and introduce U-Trans, a row-uniform full-range mixer based on node-mean broadcasting.

(2) Through controlled ablations, efficiency analysis, and public-baseline comparison, we show that global mean broadcasting is a strong global spatial baseline.

(3) We decompose attention into a row-uniform background and a non-uniform residual, and provide evidence that the residual has dataset-dependent effects, offering a plausible explanation for the absence of a consistent advantage for attention across datasets.

\section{Related Work}

\subsection{Spatial Modeling and Simple Baselines}

Traffic forecasting spatial modules range from fixed topology propagation to flexible cross-node interaction. Early models use road networks, diffusion, graph convolutions, or graph attention \citep{Guo_Lin_Feng_Song_Wan_2019,li2018diffusion,ijcai2018p0505}. Adaptive and dynamic variants use node embeddings, graph learning, continuous dynamics, and dynamic spatio-temporal graphs \citep{NEURIPS2020_ce1aad92,Choi_Choi_Hwang_Park_2022,10.1145/3447548.3467430,10.14778/3551793.3551827,ijcai2019p264,10.1145/3394486.3403118}.

Another line questions whether complex spatial modules are always necessary. Spatial-temporal identity, adaptive embeddings, MLP predictors, linear models, and lightweight baselines remain competitive when representation and training are strong \citep{ICLR2025_27c546ab,10.1145/3583780.3615160,liu2023reallyneedgraphneural,10.1145/3511808.3557702,wang2026channel,Zeng_Chen_Zhang_Xu_2023,10.1145/3583780.3614969}.

\subsection{Context and Low-Degree-of-Freedom Mixing}

Global and long-range context is modeled through global spatial-temporal networks, fully connected gates, multi-attention, spatial Transformers, dynamic long-range attention, spectral attention, scalable spatial patching, and context broadcasting \citep{ijcai2019p0317,10.1145/3690624.3709177,Jiang_Han_Zhao_Wang_2023,lin2025multiview,Oreshkin_Amini_Coyle_Coates_2021,xu2021spatialtemporaltransformernetworkstraffic,Zheng_Fan_Wang_Qi_2020}. Large-scale work exposes dense-interaction cost \citep{DBLP:journals/pvldb/HanZLTTX24,NEURIPS2025_54c9bfb0,ijcai2024p264}.

Low-degree-of-freedom mixing gives a complementary view. Non-local attention analysis, query-independent global context, fixed graph smoothing, MLP token mixing, Fourier mixing, and uniform attention show that global context can sometimes be carried by simple, nearly uniform operators rather than learned pairwise weights \citep{9022134,Hyeon-Woo_2023_ICCV,lee-thorp-etal-2022-fnet,NEURIPS2021_cba0a4ee,8578911,pmlr-v97-wu19e}.

\section{Controlled Experimental Framework}
\label{sec:framework}

\subsection{Problem Definition and Framework}

Given traffic observations over the past $L$ time steps and $N$ sensor nodes, the task is to predict the target traffic flow for the future $P$ time steps. Let the input sequence and prediction target be
\begin{equation}
\mathbf{X}_{t-L+1:t}
=
\left[\mathbf{X}_{t-L+1},\ldots,\mathbf{X}_{t}\right]
\in \mathbb{R}^{L\times N\times F}
\label{eq:input-sequence}
\end{equation}
\begin{equation}
\mathbf{Y}_{t+1:t+P}
=
\left[y_{t+1},\ldots,y_{t+P}\right]
\in \mathbb{R}^{P\times N\times C_y}
\label{eq:prediction-target}
\end{equation}
where $L$ is the historical window length, $P$ is the prediction window length, $N$ is the number of nodes, $F$ is the input feature dimension, and $C_y$ is the output channel dimension.

\begin{figure*}[t]
\centering
\includegraphics{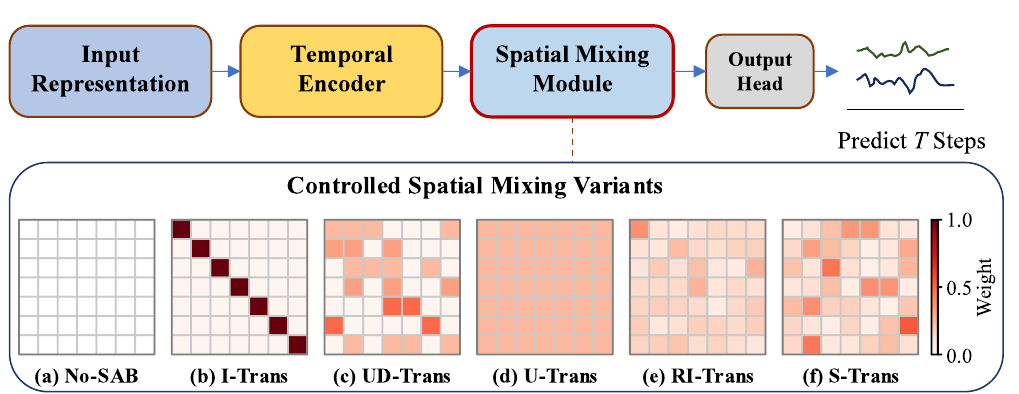}
\caption{Unified forecasting framework.}
\label{fig:framework}
\end{figure*}

The controlled framework has four stages: input representation, temporal encoding, spatial mixing, and prediction output, as shown in Figure~\ref{fig:framework}. It fixes the non-spatial components and replaces only the spatial mixing module to isolate the effect of cross-node information extraction.

\subsection{Fixed Input Representation}

All spatial variants use the same input representation, formed by concatenating projected traffic flow, congestion degree, temporal embeddings, and adaptive embeddings. Congestion degree is computed from the training-set free-flow speed; TOD/DOW settings and projection details are given in Appendix~A of the supplementary material. The resulting representation is:
\begin{equation}
\mathbf{Z}
=
\operatorname{Concat}\!\left(
\mathbf{Z}^{F},
\mathbf{Z}^{C},
\mathbf{Z}^{\mathrm{TOD}},
\mathbf{Z}^{\mathrm{DOW}},
\mathbf{E}^{\mathrm{adp}}
\right)
\label{eq:input-representation}
\end{equation}

\subsection{Fixed Temporal Encoder}

All spatial variants share the same temporal encoder. It applies standard self-attention over historical time independently for each node, followed by residual connections, LayerNorm, dropout, and a feed-forward layer. The resulting representation is passed to the spatial mixing module. Detailed equations are given in Appendix~A.

\subsection{Controlled Spatial Mixing Module}

The spatial module receives the temporally encoded multi-source node representations and compares five spatial settings: No Spatial Aggregation Block (No-SAB), Identity Transformer (I-Trans), Upstream-Downstream Transformer (UD-Trans), Uniform Transformer (U-Trans), and Standard Spatial Transformer (S-Trans). Except for No-SAB, all variants share the same residual, normalization, output-projection, and FFN scaffold. Their spatial mixers differ in the aggregation rule and in the parameterization required by that rule.
\begin{equation}
\widetilde{H}_{\tau}
=
A V_{\tau},
\qquad
V_{\tau}
=
H_{\tau}^{T}W_{V},
\qquad
A\in\mathbb{R}^{N\times N}
\label{eq:spatial-mixing}
\end{equation}

For I-Trans, the aggregation matrix is the identity:
\begin{equation}
A_{ij}^{I}=\mathbf{1}[i=j]
\label{eq:identity-mixing}
\end{equation}

For UD-Trans, the $k$-hop upstream and downstream neighborhoods of node $i$ are combined and row-averaged; the resulting local matrix is pre-normalized and directly used in the spatial block:
\begin{equation}
\mathcal{N}_{i}^{\mathrm{UD},k}
=
\mathcal{U}_{i}^{k}\cup\mathcal{D}_{i}^{k}
\label{eq:ud-neighborhood}
\end{equation}
\begin{equation}
A_{ij}^{\mathrm{UD},k}
=
\left|\mathcal{N}_{i}^{\mathrm{UD},k}\right|^{-1}
\mathbf{1}\!\left[j\in\mathcal{N}_{i}^{\mathrm{UD},k}\right]
\label{eq:ud-mixing}
\end{equation}

If this set is empty, UD-Trans falls back to self-propagation. For U-Trans, the aggregation matrix is row-uniform and is implemented by broadcasting the node mean instead of materializing an $N\times N$ matrix:
\begin{equation}
A_{ij}^{U}=\frac{1}{N}
\label{eq:uniform-mixing}
\end{equation}

S-Trans computes an input-dependent attention matrix for each head:
\begin{equation}
A_{\mathrm{attn}}^{h}
=
\operatorname{softmax}\!\left(
\frac{Q^{h}(K^{h})^{\top}}{\sqrt{d_s}}
\right)
\label{eq:spatial-attention}
\end{equation}

Only S-Trans learns input-dependent, head-specific aggregation matrices and therefore uses a multi-head formulation. I-Trans, UD-Trans, and U-Trans use single-head fixed mixing because duplicating a fixed matrix across heads adds no head-specific spatial weighting.

More specifically, let $V=[V^{(1)},\ldots,V^{(H)}]$ denote the value representation partitioned across $H$ heads in multi-head U-Trans (MH-U-Trans). When all heads use the same row-uniform matrix $U$,
\begin{equation}
\left[UV^{(1)},\ldots,UV^{(H)}\right]
=
U\left[V^{(1)},\ldots,V^{(H)}\right]
=
UV
\label{eq:multihead-uniform-equivalence}
\end{equation}

Thus, MH-U-Trans is mathematically equivalent to U-Trans under the same value and output projections. After value aggregation, the shared implementation is
\begin{equation}
O_{\tau}
=
\operatorname{Dropout}\!\left(
\widetilde{H}_{\tau}W_{O}
\right)
\label{eq:spatial-output-projection}
\end{equation}
\begin{equation}
H'_{\tau}
=
\operatorname{LayerNorm}\!\left(
H_{\tau}^{T}+O_{\tau}
\right)
\label{eq:spatial-residual}
\end{equation}
\begin{equation}
H_{\tau}^{S}
=
\operatorname{LayerNorm}\!\left(
H'_{\tau}
+
\operatorname{Dropout}\!\left(
\operatorname{FFN}(H'_{\tau})
\right)
\right)
\label{eq:spatial-ffn}
\end{equation}

No-SAB directly forwards the temporal representation to the output layer and skips the entire spatial block.

The No-SAB--I-Trans comparison provides a simple control for the non-attention components of the spatial block.

RI-Trans defines a coefficient-constrained family connecting MH-U-Trans and S-Trans. Given the head-wise attention matrix $A_{\mathrm{attn},h}$, the effective aggregation weights are
\begin{equation}
A_{\mathrm{eff},h}(r)
=
(1-r)A^{U}
+rA_{\mathrm{attn},h},
\qquad 0\le r\le 1
\label{eq:ri-trans}
\end{equation}
where $r$ controls the residual attention strength. RI-Trans becomes MH-U-Trans when $r=0$ and S-Trans when $r=1$; intermediate values retain input-dependent non-uniform attention on the uniform background. Mixing occurs after softmax and before value aggregation.

\subsection{Output Layer}

The output layer maps the representation after the spatial module to the future $P$ prediction steps. For each node, the historical temporal dimension and hidden dimension are flattened and then mapped linearly; all nodes are concatenated to obtain the final prediction:
\begin{equation}
\begin{aligned}
\widehat{\mathbf{y}}_{i}
&=
W_{\mathrm{out}}
\operatorname{Flatten}\!\left(H^{S}_{:,i,:}\right)
+b_{\mathrm{out}},\\
\widehat{\mathbf{Y}}
&\in
\mathbb{R}^{P\times N\times C_y}
\end{aligned}
\label{eq:output-layer}
\end{equation}

\section{Experiments and Results}
\label{sec:experiments}

\begin{table}[!b]
\centering
\small
\setlength{\tabcolsep}{4pt}
\begin{tabular}{lrrc}
\toprule
\textbf{Name} & \textbf{Nodes} & \textbf{Time steps} &
\textbf{Interval} \\
\midrule
PEMS03  & 358                  & 26,208 & 5 min  \\
PEMS04  & 307                  & 16,992 & 5 min  \\
PEMS07  & 883                  & 28,224 & 5 min  \\
PEMS08  & 170                  & 17,856 & 5 min  \\
T-DRIVE & 1024 ($32\times32$) & 3,600  & 60 min \\
NYCTAXI & 75 ($15\times5$)    & 17,520 & 30 min \\
\bottomrule
\end{tabular}
\normalsize
\caption{Dataset description.}
\label{tab:datasets}
\end{table}

\subsection{Experimental Settings}

\subsubsection{Datasets.}
The PEMS datasets predict total traffic flow ($C_y=1$), whereas the grid
datasets predict inflow and outflow ($C_y=2$). To test whether the
conclusion depends on sensor scale or data type, the experiments cover
four PEMS road-sensor datasets, namely PEMS03, PEMS04, PEMS07, and
PEMS08 \citep{song2020stsgcn}, and two grid traffic datasets, T-DRIVE
\citep{pan2019urban} and NYCTAXI \citep{liu2021dynamic}, with the basic
statistics collected in Table~\ref{tab:datasets}. PEMS uses 12 past
steps to predict 12 future steps; T-DRIVE and NYCTAXI use 6 past steps
to predict the next step. The chronological splits are 6:2:2 for PEMS
and 7:1:2 for the grid datasets.

\subsubsection{Implementation.}
To reduce tuning-related confounders, unless otherwise specified, our
hyperparameter settings largely follow those used in STAEformer
\citep{10.1145/3583780.3615160}, a structurally related
Transformer-based model for traffic forecasting.
All data are Z-score normalized with training-set statistics reused for
validation and test sets. Only PEMS datasets use congestion-degree
embeddings. Speed information was extracted from Caltrans PeMS
\citep{caltrans_pems}. Flow, congestion-degree, TOD, and DOW embeddings
are all 24-dimensional; the adaptive embedding is 80-dimensional. The temporal
block count is 3. Both the temporal and spatial attention modules use four heads. The spatial block count is 1 on PEMS03, PEMS04,
PEMS08, and NYCTAXI, and 3 on PEMS07 and T-DRIVE; Appendix~B reports
spatial-depth sensitivity. PEMS03 uses only TOD, whereas the other
datasets use TOD and DOW. UD-Trans uses the 5-hop upstream-downstream
neighborhood on all PEMS datasets.

\subsubsection{Training Settings.}
PEMS uses Adam, batch size 16, and early stopping with patience 10;
T-DRIVE and NYCTAXI use AdamW, batch size 16, and patience 50. All
spatial settings use five random seeds; prediction
results are reported as mean $\pm$ standard deviation. Experiments
run on an NVIDIA GeForce RTX 4090 GPU.

\subsubsection{Metrics.}
For PEMS, we report average MAE, RMSE, and MAPE over 12 prediction
steps. Because T-DRIVE and NYCTAXI contain many zero-flow grid cells, we
report masked MAE, RMSE, and MAPE with a mask threshold of 10. Inflow
and outflow metrics are computed separately and averaged. Since these
grid datasets lack directed road adjacency, UD-Trans is reported only
for PEMS.

\subsection{Main Test: Does Adaptive Attention Provide a Consistent Advantage?}

\begin{figure}[!b]
\centering
\includegraphics{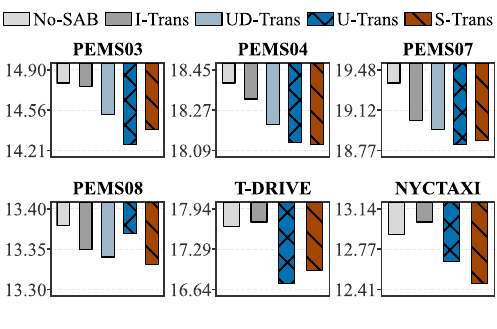}
\caption{MAE comparison on six datasets.}
\label{fig:mae-comparison}
\end{figure}

The controlled comparison includes No-SAB, I-Trans, UD-Trans, U-Trans,
and S-Trans. Figure~\ref{fig:mae-comparison} reports the MAE results,
with complete metrics provided in Table~\ref{tab:complete-results}. 

\begin{table*}[!t]
\centering
{\small
\renewcommand{\arraystretch}{1.12} 
\setlength{\tabcolsep}{2pt}
\begin{tabular*}{\textwidth}{@{\extracolsep{\fill}}llcccccc@{}}
\toprule
\textbf{Module} & \textbf{Metric} & \textbf{PEMS03} & \textbf{PEMS04} & \textbf{PEMS07} & \textbf{PEMS08} & \textbf{T-DRIVE} & \textbf{NYCTAXI} \\
\midrule
\multirow{3}{*}{No-SAB}
& MAE  & $14.79\pm0.05$ & $18.39\pm0.04$ & $19.36\pm0.04$ & $13.38\pm0.03$ & $17.66\pm0.07$ & $12.91\pm0.06$ \\
& RMSE & $25.54\pm0.08$ & $30.11\pm0.05$ & $32.62\pm0.09$ & $22.86\pm0.08$ & $31.69\pm0.21$ & $21.23\pm0.12$ \\
& MAPE & $15.11\pm0.04$ & $11.96\pm0.04$ & $8.15\pm0.03$ & $8.88\pm0.04$ & $14.76\pm0.18$ & $13.80\pm0.06$ \\
\midrule
\multirow{3}{*}{I-Trans}
& MAE  & $14.76\pm0.03$ & $18.32\pm0.04$ & $19.03\pm0.05$ & $13.35\pm0.02$ & $17.73\pm0.12$ & $13.02\pm0.07$ \\
& RMSE & $25.40\pm0.14$ & $30.12\pm0.04$ & $32.45\pm0.13$ & $22.90\pm0.11$ & $31.97\pm0.30$ & $21.37\pm0.14$ \\
& MAPE & $15.17\pm0.02$ & $11.89\pm0.06$ & $7.95\pm0.04$ & $8.88\pm0.03$ & $14.68\pm0.11$ & $13.89\pm0.10$ \\
\midrule
\multirow{3}{*}{UD-Trans}
& MAE  & $14.52\pm0.06$ & $18.21\pm0.02$ & $18.96\pm0.04$ & $13.34\pm0.03$ & --- & --- \\
& RMSE & $25.32\pm0.13$ & $30.05\pm0.04$ & $32.38\pm0.08$ & $22.81\pm0.07$ & --- & --- \\
& MAPE & $14.73\pm0.05$ & $11.84\pm0.03$ & $7.92\pm0.02$ & $8.86\pm0.03$ & --- & --- \\
\midrule
\multirow{3}{*}{U-Trans}
& MAE  & $14.26\pm0.04$ & $18.13\pm0.03$ & $18.82\pm0.03$ & $13.37\pm0.03$ & $16.73\pm0.13$ & $12.66\pm0.07$ \\
& RMSE & $24.88\pm0.08$ & $30.01\pm0.08$ & $32.34\pm0.07$ & $22.94\pm0.13$ & $30.57\pm0.26$ & $20.92\pm0.11$ \\
& MAPE & $14.54\pm0.05$ & $11.78\pm0.01$ & $7.86\pm0.01$ & $8.85\pm0.03$ & $13.93\pm0.08$ & $13.37\pm0.10$ \\
\midrule
\multirow{3}{*}{S-Trans}
& MAE  & $14.39\pm0.03$ & $18.12\pm0.02$ & $18.86\pm0.07$ & $13.33\pm0.06$ & $16.94\pm0.22$ & $12.46\pm0.12$ \\
& RMSE & $25.99\pm0.14$ & $30.03\pm0.04$ & $32.36\pm0.14$ & $22.87\pm0.11$ & $30.70\pm0.18$ & $20.60\pm0.23$ \\
& MAPE & $14.79\pm0.04$ & $11.80\pm0.03$ & $7.89\pm0.04$ & $8.85\pm0.06$ & $14.16\pm0.25$ & $13.26\pm0.20$ \\
\bottomrule
\end{tabular*}
}
\caption{Controlled spatial-mixer results on six datasets (mean $\pm$ std over five seeds).}
\label{tab:complete-results}
\end{table*}

Before comparing aggregation rules, we first examine No-SAB and
I-Trans. Their MAE differences are small and vary in direction across
datasets, indicating that the non-attention components of the spatial
block alone do not provide a consistent forecasting gain.

U-Trans achieves lower mean MAE than No-SAB on all six
datasets. More importantly, I-Trans and U-Trans share the same
spatial-block scaffold and differ only in node aggregation: I-Trans uses
identity aggregation, whereas U-Trans broadcasts the full-range node
mean. U-Trans achieves lower MAE than I-Trans on five of the six
datasets, suggesting that row-uniform full-range propagation
usually provides predictive gains beyond node-wise transformation
alone. On the four PEMS datasets, U-Trans also outperforms or closely
matches UD-Trans, showing that full-range uniform mixing is competitive
with topology-constrained local aggregation. Across the six datasets, S-Trans yields relative MAE changes of \(-1.58\%\) to \(+1.26\%\) compared with U-Trans. Within this controlled framework, S-Trans does not consistently outperform U-Trans. Table~\ref{tab:complete-results} shows the same cross-dataset finding for RMSE and MAPE.

Taken together, these controls distinguish the value of global propagation from that of adaptive pairwise weighting: the former is generally useful, whereas the latter provides no consistent additional benefit under the evaluated settings.

\subsubsection{Robustness to Input Representations.}
The main experiments use a multi-source input representation with a high-dimensional adaptive embedding. We therefore further vary the input representations across all six datasets by removing auxiliary embeddings in turn and reducing the dimensionality of the adaptive embedding. S-Trans exhibits a consistent cross-dataset advantage only under the flow-only setting. After auxiliary representation branches are introduced, this advantage no longer persists across datasets, suggesting that richer input representations may reduce the marginal benefit of adaptive attention on some datasets. TOD/DOW information is directly available from timestamps, whereas adaptive embeddings are learned end-to-end. Neither type of branch requires additional exogenous data, and richer input configurations incorporating them substantially reduce prediction error relative to the flow-only setting. Therefore, for U-Trans and S-Trans under the evaluated settings, flow-only input is not the best-performing configuration. Under a fixed computational budget, U-Trans can exploit the lower complexity of its spatial mixer by reallocating the saved computation to additional or higher-dimensional embeddings. Complete results are provided in Appendix~C.

\subsection{Efficiency of Global Mean Broadcasting}

Efficiency is evaluated on PEMS08, PEMS04, and PEMS07 to cover small,
medium, and large node scales. Table~\ref{tab:efficiency} reports
per-epoch training time and test inference time, with node count
$N$ and spatial-layer count $L_s$.

\begin{table}[!t]
\centering
\setlength{\tabcolsep}{1mm}
\begin{tabular}{lcccccc}
\toprule
\multicolumn{1}{c}{
  \shortstack[c]{\strut\\\textbf{Dataset}\\\strut}
}
& \multicolumn{2}{c}{
  \shortstack{\textbf{PEMS08}\\$N=170$\\$L_s=1$}
}
& \multicolumn{2}{c}{
  \shortstack{\textbf{PEMS04}\\$N=307$\\$L_s=1$}
}
& \multicolumn{2}{c}{
  \shortstack{\textbf{PEMS07}\\$N=883$\\$L_s=3$}
} \\
\cmidrule(lr){2-3}
\cmidrule(lr){4-5}
\cmidrule(lr){6-7}
\textbf{Model}
& \textbf{Train} & \textbf{Infer}
& \textbf{Train} & \textbf{Infer}
& \textbf{Train} & \textbf{Infer} \\
\midrule
No-SAB  & 33.0 & 1.5 & 30.7 & 2.4 & 81.2  & 9.9  \\
U-Trans & 36.1 & 1.8 & 33.4 & 2.8 & 126.8 & 15.3 \\
S-Trans & 37.9 & 2.1 & 39.6 & 3.7 & 352.5 & 51.1 \\
\bottomrule
\end{tabular}
\caption{Efficiency comparison. Unit: seconds.}
\label{tab:efficiency}
\end{table}

U-Trans is consistently faster than S-Trans, and the gap widens with
node scale and spatial depth. On PEMS07, S-Trans is 225.7 s slower per
training epoch and 35.8 s slower in full-test inference. This matches
the implementation: U-Trans computes and broadcasts the node mean in
$O(N)$, whereas S-Trans forms dense node-to-node attention weights in
$O(N^2)$.

\newcommand{\TableFourRowStretch}{1.31}
\begin{table*}[!t]
\centering
\small
\renewcommand{\arraystretch}{\TableFourRowStretch}
\setlength{\tabcolsep}{1mm}
\begin{tabular*}{\textwidth}{
  @{\extracolsep{\fill}}llccccccccc@{}
}
\toprule
\textbf{Dataset} & \textbf{Metric}
& \makebox[1.42cm][c]{\scriptsize\textbf{GWNet}}
& \makebox[1.42cm][c]{\scriptsize\textbf{STGNCDE}}
& \makebox[1.42cm][c]{\scriptsize\textbf{ASTGNN}}
& \makebox[1.42cm][c]{\scriptsize\textbf{PDFormer}}
& \makebox[1.42cm][c]{\scriptsize\textbf{STAEformer}}
& \makebox[1.42cm][c]{\scriptsize\textbf{STD-PLM}}
& \makebox[1.42cm][c]{\scriptsize\textbf{STSGNN}}
& \makebox[1.42cm][c]{\scriptsize\textbf{MAGE}}
& \makebox[1.42cm][c]{\scriptsize\textbf{U-Trans}} \\
\midrule
        & MAE
& \makebox[1.42cm][c]{\underline{14.59}}
& \makebox[1.42cm][c]{15.57}
& \makebox[1.42cm][c]{14.78}
& \makebox[1.42cm][c]{14.94}
& \makebox[1.42cm][c]{15.35}
& \makebox[1.42cm][c]{\underline{14.59}}
& \makebox[1.42cm][c]{14.61}
& \makebox[1.42cm][c]{14.72}
& \makebox[1.42cm][c]{\textbf{14.26}} \\
PEMS03 & MAPE
& \makebox[1.42cm][c]{15.52}
& \makebox[1.42cm][c]{15.06}
& \makebox[1.42cm][c]{14.79}
& \makebox[1.42cm][c]{15.82}
& \makebox[1.42cm][c]{15.18}
& \makebox[1.42cm][c]{14.92}
& \makebox[1.42cm][c]{\textbf{14.43}}
& \makebox[1.42cm][c]{14.87}
& \makebox[1.42cm][c]{\underline{14.54}} \\
        & RMSE
& \makebox[1.42cm][c]{25.24}
& \makebox[1.42cm][c]{27.09}
& \makebox[1.42cm][c]{25.00}
& \makebox[1.42cm][c]{25.39}
& \makebox[1.42cm][c]{27.55}
& \makebox[1.42cm][c]{25.36}
& \makebox[1.42cm][c]{25.24}
& \makebox[1.42cm][c]{\textbf{23.73}}
& \makebox[1.42cm][c]{\underline{24.88}} \\
\cmidrule(lr){2-11}
        & MAE
& \makebox[1.42cm][c]{18.53}
& \makebox[1.42cm][c]{19.21}
& \makebox[1.42cm][c]{18.60}
& \makebox[1.42cm][c]{18.32}
& \makebox[1.42cm][c]{18.22}
& \makebox[1.42cm][c]{\underline{18.16}}
& \makebox[1.42cm][c]{18.69}
& \makebox[1.42cm][c]{\underline{18.16}}
& \makebox[1.42cm][c]{\textbf{18.13}} \\
PEMS04 & MAPE
& \makebox[1.42cm][c]{12.89}
& \makebox[1.42cm][c]{12.76}
& \makebox[1.42cm][c]{12.36}
& \makebox[1.42cm][c]{12.10}
& \makebox[1.42cm][c]{11.98}
& \makebox[1.42cm][c]{\underline{11.89}}
& \makebox[1.42cm][c]{12.24}
& \makebox[1.42cm][c]{12.64}
& \makebox[1.42cm][c]{\textbf{11.78}} \\
        & RMSE
& \makebox[1.42cm][c]{\textbf{29.92}}
& \makebox[1.42cm][c]{31.09}
& \makebox[1.42cm][c]{30.91}
& \makebox[1.42cm][c]{\underline{29.97}}
& \makebox[1.42cm][c]{30.18}
& \makebox[1.42cm][c]{30.21}
& \makebox[1.42cm][c]{30.84}
& \makebox[1.42cm][c]{30.16}
& \makebox[1.42cm][c]{30.01} \\
\cmidrule(lr){2-11}
        & MAE
& \makebox[1.42cm][c]{20.47}
& \makebox[1.42cm][c]{20.53}
& \makebox[1.42cm][c]{20.62}
& \makebox[1.42cm][c]{19.83}
& \makebox[1.42cm][c]{\underline{19.14}}
& \makebox[1.42cm][c]{19.25}
& \makebox[1.42cm][c]{19.87}
& \makebox[1.42cm][c]{19.49}
& \makebox[1.42cm][c]{\textbf{18.82}} \\
PEMS07 & MAPE
& \makebox[1.42cm][c]{8.61}
& \makebox[1.42cm][c]{8.80}
& \makebox[1.42cm][c]{8.86}
& \makebox[1.42cm][c]{8.53}
& \makebox[1.42cm][c]{\underline{8.01}}
& \makebox[1.42cm][c]{8.06}
& \makebox[1.42cm][c]{8.29}
& \makebox[1.42cm][c]{8.25}
& \makebox[1.42cm][c]{\textbf{7.86}} \\
        & RMSE
& \makebox[1.42cm][c]{33.47}
& \makebox[1.42cm][c]{33.84}
& \makebox[1.42cm][c]{34.00}
& \makebox[1.42cm][c]{32.87}
& \makebox[1.42cm][c]{32.60}
& \makebox[1.42cm][c]{32.84}
& \makebox[1.42cm][c]{32.97}
& \makebox[1.42cm][c]{\underline{32.50}}
& \makebox[1.42cm][c]{\textbf{32.34}} \\
\cmidrule(lr){2-11}
        & MAE
& \makebox[1.42cm][c]{14.40}
& \makebox[1.42cm][c]{15.45}
& \makebox[1.42cm][c]{15.00}
& \makebox[1.42cm][c]{13.58}
& \makebox[1.42cm][c]{13.46}
& \makebox[1.42cm][c]{\textbf{13.31}}
& \makebox[1.42cm][c]{14.38}
& \makebox[1.42cm][c]{13.66}
& \makebox[1.42cm][c]{\underline{13.37}} \\
PEMS08 & MAPE
& \makebox[1.42cm][c]{9.21}
& \makebox[1.42cm][c]{9.92}
& \makebox[1.42cm][c]{9.50}
& \makebox[1.42cm][c]{9.05}
& \makebox[1.42cm][c]{8.88}
& \makebox[1.42cm][c]{\textbf{8.84}}
& \makebox[1.42cm][c]{9.25}
& \makebox[1.42cm][c]{9.09}
& \makebox[1.42cm][c]{\underline{8.85}} \\
        & RMSE
& \makebox[1.42cm][c]{23.39}
& \makebox[1.42cm][c]{24.81}
& \makebox[1.42cm][c]{24.70}
& \makebox[1.42cm][c]{23.51}
& \makebox[1.42cm][c]{23.25}
& \makebox[1.42cm][c]{23.19}
& \makebox[1.42cm][c]{23.81}
& \makebox[1.42cm][c]{\underline{23.04}}
& \makebox[1.42cm][c]{\textbf{22.94}} \\
\bottomrule
\end{tabular*}
\normalsize
\caption{PEMS baseline comparison. For
Tables~\ref{tab:pems-baselines} and~\ref{tab:grid-baselines}, bold and
underlined values indicate best and second-best results.}
\label{tab:pems-baselines}
\end{table*}

\begin{table}[!t]
\centering
\small
\setlength{\tabcolsep}{1pt}
\begin{tabular}{lcccccc}
\toprule
\multicolumn{1}{c}{\textbf{Dataset}}
& \multicolumn{3}{c}{\textbf{T-DRIVE}}
& \multicolumn{3}{c}{\textbf{NYCTAXI}} \\
\cmidrule(lr){2-4}
\cmidrule(lr){5-7}
\multicolumn{1}{c}{\textbf{Metric}}
& \textbf{MAE} & \textbf{RMSE} & \textbf{MAPE}
& \textbf{MAE} & \textbf{RMSE} & \textbf{MAPE} \\
\midrule
GWNet
& 19.55 & 36.18 & 16.56
& 13.30 & 21.71 & 13.94 \\
STGNCDE
& 19.29 & 36.12 & 16.50
& 13.28 & 21.68 & 13.93 \\
ASTGNN
& 18.79 & 33.93 & 15.84
& 12.98 & 21.19 & 13.65 \\
PDFormer
& 17.79 & 31.55 & 14.68
& \textbf{12.36} & \textbf{20.18} & \textbf{12.78} \\
STAEformer
& \underline{16.97} & 31.02 & \underline{13.81}
& 12.61 & 20.53 & \underline{12.96} \\
Cy2Mixer
& 16.99 & \underline{30.82} & \textbf{13.56}
& \underline{12.59} & \underline{20.45} & 13.03 \\
U-Trans
& \textbf{16.73} & \textbf{30.57} & 13.93
& 12.66 & 20.92 & 13.37 \\
\bottomrule
\end{tabular}
\normalsize
\caption{Grid public-baseline comparison. Baseline results are from
PDFormer and Cy2Mixer.}
\label{tab:grid-baselines}
\end{table}

\FloatBarrier

\subsection{Public-Baseline Comparison}

\subsubsection{Baselines.}
We compare U-Trans with the following nine public baselines, belonging
to three categories. 

(1) \textbf{\textit{Graph neural network-based
models:}} GWNet \citep{ijcai2019p264}, STGNCDE
\citep{Choi_Choi_Hwang_Park_2022}, STSGNN
\citep{Chen_Lin_Huo_Yan_2025}, MAGE
\citep{NEURIPS2025_54c9bfb0}, and Cy2Mixer
\citep{pmlr-v269-lee25a}. 

(2) \textbf{\textit{Self-attention/Transformer-based
models:}} ASTGNN \citep{9346058} and PDFormer
\citep{Jiang_Han_Zhao_Wang_2023}. 

(3)
\textbf{\textit{Representation-centric and PLM-based models:}}
STAEformer \citep{10.1145/3583780.3615160} and STD-PLM
\citep{Huang_Mao_Guo_Chen_Shen_Li_Lin_Wan_2025}.

Tables~\ref{tab:pems-baselines} and~\ref{tab:grid-baselines} report
PEMS and grid results, respectively, under each dataset's standard
setting. Baseline results are collected from published papers and
comparison tables; U-Trans results are from our experiments.

\subsubsection{Results.}
On PEMS datasets, U-Trans achieves the lowest reported MAE on PEMS03,
PEMS04, and PEMS07, remains close to the lowest reported MAE on PEMS08,
and obtains the best or near-best reported MAPE/RMSE in most cases. On
grid datasets, it gives the lowest reported MAE and RMSE on T-DRIVE and
remains competitive on NYCTAXI. These results indicate that row-uniform
global mean broadcasting is also competitive against graph-based and
Transformer-based public baselines.

\section[Mechanism Analysis: Uniform Global Background and Non-Uniform Attention Residuals]{Mechanism Analysis: Uniform Global Background and Non-Uniform Attention Residuals}
\label{sec:mechanism-analysis}

We examine this pattern by decomposing S-Trans attention into a row-uniform global background and a non-uniform residual, and then testing the residual through interpolation and front global-mean injection. Based on the six-dataset endpoint results, we select NYCTAXI, PEMS03, and PEMS04 to cover beneficial, detrimental, and nearly neutral residual effects. These diagnostic cases characterize how the sign and magnitude of the residual effect vary across datasets.

\subsection{Uniform Global Background and Attention-Residual Decomposition}
\label{subsec:uniform-background-residual}

We hypothesize that the value-projected representation of node $j$ at time $t$ in the input to the spatial module can be decomposed into a latent shared global state and a node-local fluctuation:
\begin{equation}
 v_{j,t}=g_t+\epsilon_{j,t}
\label{eq:latent-global-state}
\end{equation}
where $g_t$ denotes a hypothesized latent shared global state, including the overall demand level, network-wide congestion background, and shared temporal trend, and $\epsilon_{j,t}$ denotes node-specific local fluctuations around this shared background, including regional heterogeneity and short-term localized dynamics. The global state $g_t$ is not the arithmetic node mean. The U-Trans mean term in the prediction model is an empirical proxy for it.

Under this decomposition, the uniform global message of U-Trans can be written as:
\begin{equation}
 m_t^{U-Trans}
 =\frac{1}{N}\sum_{j=1}^{N}v_{j,t}
 =g_t+\frac{1}{N}\sum_{j=1}^{N}\epsilon_{j,t}
\label{eq:uniform-global-message}
\end{equation}
Equation~\eqref{eq:uniform-global-message} provides a mean-based approximation of the shared background. With representative node coverage, averaging suppresses node-local fluctuations while preserving the major trend; sparse coverage or strong regional heterogeneity can still bias this approximation. Since the spatial block is residual, U-Trans adds this smoothed background without overwriting node representations.

S-Trans can be written as a multi-head row-uniform attention background plus an input-dependent non-uniform attention residual. Let the row-uniform weight matrix be $U$, where $U_{ij}=1/N$, and let $\mathbf{1}\in\mathbb{R}^{N}$ denote the all-ones vector; the standard node-wise attention matrix can be decomposed as:
\begin{equation}
 A_t=U+\Delta_t,
 \qquad
 U=\frac{1}{N}\mathbf{1}\mathbf{1}^{\top},
 \qquad
 \Delta_t\mathbf{1}=0
\label{eq:attention-decomposition}
\end{equation}
In Equation~\eqref{eq:attention-decomposition}, $\Delta_t$ denotes the deviation of attention weights from the uniform distribution. Because the sum of each row of the attention weights after softmax is 1, and the sum of each row of the uniform matrix $U$ is also 1, the deviation term satisfies $\Delta_t\mathbf{1}=0$. Thus, the S-Trans spatial message of the $i$-th node can be written as:
\begin{equation}
 m_{i,t}^{S-Trans}
 =\sum_{j=1}^{N}A_{ij,t}v_{j,t}
 =m_t^U+\sum_{j=1}^{N}\Delta_{ij,t}v_{j,t}
\label{eq:strans-spatial-message}
\end{equation}
In Equation~\eqref{eq:strans-spatial-message}, $m_t^U=\sum_{j=1}^{N}U_{ij}v_{j,t}$ denotes the row-uniform attention background; the multi-head case applies the same decomposition head-wise.

Further substituting Equation~\eqref{eq:latent-global-state} gives:
\begin{equation}
 \sum_{j=1}^{N}\Delta_{ij,t}v_{j,t}
 =\sum_{j=1}^{N}\Delta_{ij,t}\left(g_t+\epsilon_{j,t}\right)
 =\sum_{j=1}^{N}\Delta_{ij,t}\epsilon_{j,t}
\label{eq:residual-cancels-global-state}
\end{equation}
Equation~\eqref{eq:residual-cancels-global-state} shows that the attention residual cancels the common-state component because each deviation row sums to zero. Under the proposed decomposition, the non-uniform residual operates on node-specific fluctuations and does not provide a stronger estimate of the shared global state.

Thus, the residual value depends on whether node-local structures are predictive: it can reduce, increase, or barely change MAE relative to MH-U-Trans. Attention entropy in the subsequent residual-strength experiments further assesses the empirical relevance of this decomposition.

\subsection{Residual Diagnostics and Front Global-Mean Injection}
\label{subsec:residual-diagnostics}

Before independently optimizing RI-Trans, we first perform an inference-only interpolation by applying Equation~\eqref{eq:ri-trans} to the post-softmax attention matrices produced by the trained S-Trans checkpoints, while keeping all learned parameters fixed. At the evaluated coefficients, the five-seed mean MAE increases monotonically as $r$ decreases on the three diagnostic datasets (Appendix E). This control characterizes checkpoint sensitivity to post-training modification of the spatial mixer rather than performance after re-optimization. Its r=0 endpoint is therefore not an independently trained MH-U-Trans model. Each modified architecture in the two diagnostics below is trained independently under the same S-Trans protocol and five random seeds.

\subsubsection{(1) Residual-Strength Interpolation.}

RI-Trans controls the non-uniform attention residual through coefficient $r$. The effective attention is written as a row-uniform component plus a scaled deviation:
\begin{equation}
 A_{r,t}
 =U+r\left(A_t^{(r)}-U\right)
 =U+r\Delta_t^{(r)}
\label{eq:residual-strength-interpolation}
\end{equation}
The corresponding spatial output is:
\begin{equation}
 O_{r,t}
 =A_{r,t}V_t^{(r)}
 =UV_t^{(r)}+r\Delta_t^{(r)}V_t^{(r)}
\label{eq:residual-strength-output}
\end{equation}
Here, the attention matrix, residual, and value matrix are learned
independently for each residual coefficient $r$; the decomposition applies
head-wise. Equation~\eqref{eq:residual-strength-output} links this experiment
to the preceding attention-residual decomposition: the first term is the
row-uniform global-attention background, and the second term is the
non-uniform residual scaled by $r$. Since each deviation term has zero row
sum, the coefficient $r$ scales only the non-uniform node-to-node
recombination term and does not add a common global background. To
characterize the resulting attention distribution, normalized row entropy
is computed separately for each of the four heads and then averaged across
heads and five random seeds.

\subsubsection{(2) Front Global-Mean Injection.}

Front global-mean injection tests how direct global-mean information changes S-Trans output. For clarity, projection biases are omitted in this derivation. The row-uniform full-range mean is inserted into the node representation before S-Trans attention, increasing the row-uniform background and producing a compensated residual through changed query-key geometry. The compensated input adds this message before query, key, and value projection:
\begin{equation}
 H_t'=H_t+UH_t=(I+U)H_t
\label{eq:front-mean-input}
\end{equation}
Because value projection is linear, the compensated value matrix becomes:
\begin{equation}
V_t'=H_t'W_V=(I+U)V_t=V_t+UV_t
\label{eq:front-mean-value}
\end{equation}
Let the attention matrix produced from the compensated representation be decomposed as:
\begin{equation}
 A_t'=U+\Delta_t',
 \qquad
 \Delta_t'\mathbf{1}=0
\label{eq:compensated-attention-decomposition}
\end{equation}
The compensated S-Trans output is therefore:
\begin{equation}
 O_{pre,t}
 =A_t'V_t'
 =A_t'V_t+UV_t
\label{eq:compensated-output}
\end{equation}
Using the row-stochastic property of attention, namely $A_t'\mathbf{1}=\mathbf{1}$ and hence $A_t'U=U$, the output decomposition is:
\begin{equation}
\begin{aligned}
 O_{pre,t}
 &=A_t'V_t+UV_t,\\
 A_t'V_t+UV_t
 &=\left(U+\Delta_t'\right)V_t+UV_t,\\
 \left(U+\Delta_t'\right)V_t+UV_t
 &=2UV_t+\Delta_t'V_t
\end{aligned}
\label{eq:front-compensation-decomposition}
\end{equation}
Equation~\eqref{eq:front-compensation-decomposition} shows that front compensation strengthens the row-uniform background and replaces the original residual with a compensated residual. This attention-aggregation view explains why global-mean injection centers the output more strongly on the global background while leaving only a dataset-dependent non-uniform term.

The logit-level reason for the compensated residual is summarized in Appendix D.

For both diagnostics, relative MAE change uses MH-U-Trans as the zero reference and is computed as Equation~\eqref{eq:relative-mae-change}:
\begin{equation}
 \Delta MAE(r)
 =\frac{MAE(r)-MAE(0)}{MAE(0)}\times 100\%
\label{eq:relative-mae-change}
\end{equation}

\begin{figure}[t]
\centering
\includegraphics{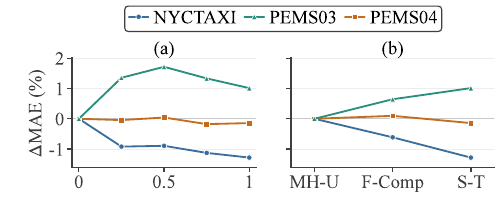}
\caption{Residual-strength interpolation over $r$ and front compensation. MH-U, F-Comp, and S-T denote MH-U-Trans, front-compensated S-Trans, and S-Trans, respectively.}
\label{fig:residual-diagnostics}
\end{figure}

Figure~\ref{fig:residual-diagnostics} visualizes the two diagnostics under the same MH-U-Trans reference. Figure~\ref{fig:residual-diagnostics}(a) reports residual-strength interpolation, and Figure~\ref{fig:residual-diagnostics}(b) reports front global-mean injection after strengthening the uniform background; complete numerical results are provided in Appendix E.

On NYCTAXI, PEMS03, and PEMS04, decreasing attention entropy as $r$ increases confirms that the interpolation progressively strengthens the non-uniform residual. The normalized head-averaged entropy decreases as 1.00/0.98/0.95/0.93/0.89 on NYCTAXI, 1.00/0.95/0.92/0.88/0.82 on PEMS03, and 1.00/0.98/0.96/0.95/0.91 on PEMS04. Even at $r=1$, the endpoint values remain relatively high, indicating substantial spatial dispersion rather than highly concentrated attention. Figure~\ref{fig:residual-diagnostics}(a) further shows that this monotonic entropy reduction does not yield monotonic forecasting improvements: all nonzero residual settings improve over $r=0$ on NYCTAXI, PEMS03 degrades most at an intermediate strength, and PEMS04 changes only marginally. At $r=1$, relative MAE changes are $+1.01\%$ on PEMS03, $-1.28\%$ on NYCTAXI, and $-0.14\%$ on PEMS04. Figure~\ref{fig:residual-diagnostics}(b) shows that the front-compensated variant yields relative MAE changes closer to zero than S-Trans, indicating performance closer to the row-uniform MH-U-Trans baseline. Together, the two diagnostics indicate that the non-uniform attention residual is dataset-dependent: it can be beneficial, detrimental, or nearly neutral. Front global-mean injection strengthens the row-uniform background and brings performance closer to MH-U-Trans; although it changes residual formation, the compensated residual remains dataset-dependent.

\section{Conclusion}
\label{sec:conclusion}

Under a controlled framework that changes only the spatial mixing module, we do not observe a consistent advantage of S-Trans over U-Trans across six road-sensor and grid-demand benchmarks, while U-Trans reduces node-scale spatial-mixing complexity from \(O(N^2)\) to \(O(N)\). Prediction ablations show that full-range mean broadcasting is a strong spatial baseline, and embedding ablations across all six benchmarks show the same cross-dataset pattern under the full and partially ablated configurations considered in this work. 

The mechanism analysis suggests that the additional capacity of spatial attention lies in its non-uniform residual beyond a row-uniform global background; its predictive value varies across datasets and may be beneficial, detrimental, or nearly neutral. Because the relative performance of U-Trans and S-Trans cannot be assumed for a target dataset, U-Trans provides a strong global spatial baseline and a simpler, more efficient starting point for evaluation.

\bibliographystyle{plainnat}
\bibliography{references}

\onecolumn

\begin{center}
  \vspace*{0.18in}
  {\LARGE\bfseries Supplementary Material\par}
  \vspace{0.08in}
  {\Large\bfseries
  Do We Really Need Adaptive Global Spatial Attention for Traffic Forecasting?\par}
\end{center}
\vspace{0.12in}

\setcounter{table}{5}

\section{Appendix A Fixed Representation and Temporal Encoder Details}

This appendix contains the detailed definitions and equations for the fixed input representation and temporal encoder summarized in the \emph{Fixed Input Representation} and \emph{Fixed Temporal Encoder} sections of the main paper. These components are kept unchanged across all spatial variants.

\subsection{A.1 Flow and Congestion-Degree Inputs}

Let the raw flow and speed of node $i$ at time $t$ be $f_{i,t}$ and $v_{i,t}$, respectively. The node-level free-flow speed is estimated on the training time set:
\begin{equation}
  v_i^{\mathrm{free}}
  = \operatorname{Quantile}_{0.95}
  \left(\left\{v_{i,t}\mid t\in\mathcal{T}_{\mathrm{train}}\right\}\right).
  \tag{A1}
\end{equation}
The congestion degree is the decline ratio of the current speed relative to the free-flow speed:
\begin{equation}
  C_{i,t}=\operatorname{clip}\left(1-\frac{v_{i,t}}{v_i^{\mathrm{free}}},0,1\right).
  \tag{A2}
\end{equation}
A larger $C_{i,t}$ indicates a stronger speed decline and more severe congestion. Let the historical flow tensor and congestion-degree tensor be $\mathbf{X}^{F}$ and $\mathbf{X}^{C}$, respectively; both are projected into the hidden space through fully connected layers:
\begin{align}
  \mathbf{Z}^{F} &= \operatorname{FC}_{F}\!\left(\mathbf{X}^{F}\right)
  \in \mathbb{R}^{L\times N\times d_F}, \tag{A3}\\
  \mathbf{Z}^{C} &= \operatorname{FC}_{C}\!\left(\mathbf{X}^{C}\right)
  \in \mathbb{R}^{L\times N\times d_C}. \tag{A4}
\end{align}

\subsection{A.2 Time Embeddings}

TOD and DOW embeddings are obtained from learnable lookup dictionaries. Let the sampling interval be $\Delta$ minutes; then one day contains $N_T=1440/\Delta$ time-of-day slots. In experiments, PEMS uses $N_T=288$, T-DRIVE uses $N_T=24$, and NYCTAXI uses $N_T=48$. Let the TOD and DOW embedding dictionaries be
\begin{equation}
  \mathbf{E}^{\mathrm{TOD}}\in\mathbb{R}^{N_T\times d_{\mathrm{tod}}},
  \qquad
  \mathbf{E}^{\mathrm{DOW}}\in\mathbb{R}^{7\times d_{\mathrm{dow}}}.
  \tag{A5}
\end{equation}
For time step $t$ within the input window, the TOD slot $r_t$ and DOW index $d_t$ select the time embeddings:
\begin{equation}
  \mathbf{z}_{t}^{\mathrm{TOD}}=\mathbf{E}_{r_t}^{\mathrm{TOD}},
  \qquad
  \mathbf{z}_{t}^{\mathrm{DOW}}=\mathbf{E}_{d_t}^{\mathrm{DOW}}.
  \tag{A6}
\end{equation}
The selected TOD and DOW embeddings are broadcast along the node dimension to form $\mathbf{Z}^{\mathrm{TOD}}\in\mathbb{R}^{L\times N\times d_{\mathrm{tod}}}$ and $\mathbf{Z}^{\mathrm{DOW}}\in\mathbb{R}^{L\times N\times d_{\mathrm{dow}}}$. Because the experimental period of PEMS03 changes markedly, only TOD is used for PEMS03; the remaining datasets use both TOD and DOW.

\subsection{A.3 Adaptive Embedding and Overall Input Representation}

The framework also includes learnable adaptive embeddings for time-step and node-related representations:
\begin{equation}
  \mathbf{E}^{\mathrm{adp}}\in\mathbb{R}^{L\times N\times d_{\mathrm{adp}}}.
  \tag{A7}
\end{equation}
All active branches are concatenated along the feature dimension to form the final input representation:
\begin{align}
  \mathbf{Z}
  &=\operatorname{Concat}\!\left(
  \mathbf{Z}^{F},\mathbf{Z}^{C},\mathbf{Z}^{\mathrm{TOD}},
  \mathbf{Z}^{\mathrm{DOW}},\mathbf{E}^{\mathrm{adp}}
  \right), \tag{A8}\\
  \mathbf{Z}&\in\mathbb{R}^{L\times N\times d},
  \qquad
  d=d_F+d_C+d_{\mathrm{tod}}+d_{\mathrm{dow}}+d_{\mathrm{adp}}. \tag{A9}
\end{align}
Unused branches have embedding dimension 0.

\subsection{A.4 Temporal Encoding Layer}

The temporal encoding layer models the historical state of each node through self-attention along the temporal dimension. Let the input of layer $\ell$ be $\mathbf{H}^{(\ell)}$. For any node $i$, the temporal query, key, and value are computed as
\begin{equation}
  \mathbf{Q}_{i}=\mathbf{H}_{:,i,:}^{(\ell)}\mathbf{W}_{Q}^{T},
  \quad
  \mathbf{K}_{i}=\mathbf{H}_{:,i,:}^{(\ell)}\mathbf{W}_{K}^{T},
  \quad
  \mathbf{V}_{i}=\mathbf{H}_{:,i,:}^{(\ell)}\mathbf{W}_{V}^{T}.
  \tag{A10}
\end{equation}
The temporal self-attention is
\begin{equation}
  \operatorname{Attn}_{T}\!\left(\mathbf{H}_{:,i,:}^{(\ell)}\right)
  =\operatorname{softmax}\!\left(
  \frac{\mathbf{Q}_{i}\mathbf{K}_{i}^{\top}}{\sqrt{d_h}}
  \right)\mathbf{V}_{i}.
  \tag{A11}
\end{equation}
where $d_h$ is the single-head attention dimension. After residual connection, LayerNorm, Dropout, and position-wise FFN, temporal blocks continue to be stacked; after $L_T$ temporal encoding layers, we obtain
\begin{equation}
  \mathbf{H}^{T}=\mathcal{T}_{\mathrm{enc}}\!\left(\mathbf{Z}\right)
  \in\mathbb{R}^{L\times N\times d}.
  \tag{A12}
\end{equation}

\section{Appendix B Spatial-Layer Ablation Experiments}

Spatial-layer ablations on PEMS04 and PEMS07 vary the number of spatial blocks from 1 to 3 while fixing the input representation, temporal encoder, output layer, optimizer, data split, and random seeds. Table 6 reports the three evaluation metrics used in the main paper: RMSE, MAE, and MAPE.

Table 6 reports the U-Trans and S-Trans results across spatial depths 1--3. On PEMS04, U-Trans MAE decreases from 18.1280 to 18.1157, whereas S-Trans increases from 18.1241 to 18.1818. On PEMS07, both improve with depth: U-Trans from 19.0069 to 18.8217 and S-Trans from 19.0792 to 18.8639. These results show that the main U-Trans--S-Trans comparison is not specific to a single spatial-depth setting.

\FloatBarrier
\begin{table}[!ht]
\centering
{\normalsize
\setlength{\tabcolsep}{3pt}
\begin{tabular*}{\textwidth}{@{\extracolsep{\fill}}llcccccc@{}}
\toprule
\multirow{2}{*}{$L_S$}
& \multirow{2}{*}{\textbf{Module}}
& \multicolumn{3}{c}{\textbf{PEMS04}}
& \multicolumn{3}{c}{\textbf{PEMS07}} \\
\cmidrule(lr){3-5}
\cmidrule(lr){6-8}
& & \textbf{MAE} & \textbf{RMSE} & \textbf{MAPE}
& \textbf{MAE} & \textbf{RMSE} & \textbf{MAPE} \\
\midrule
1 & U-Trans & 18.1280 & 30.0134 & 11.7797 & 19.0069 & 32.5545 & 7.9412 \\
1 & S-Trans & 18.1241 & 30.0293 & 11.7980 & 19.0792 & 32.5404 & 8.0342 \\
2 & U-Trans & 18.1165 & 29.9953 & 11.7664 & 18.8742 & 32.3804 & 7.8885 \\
2 & S-Trans & 18.1793 & 30.1184 & 11.8295 & 18.9186 & 32.4006 & 7.9360 \\
3 & U-Trans & 18.1157 & 29.9905 & 11.7589 & 18.8217 & 32.3368 & 7.8570 \\
3 & S-Trans & 18.1818 & 30.1416 & 11.8475 & 18.8639 & 32.3599 & 7.8889 \\
\bottomrule
\end{tabular*}
}
\caption{Spatial-layer ablation on PEMS04 and PEMS07.}
\label{tab:spatial-depth-pems04}
\label{tab:spatial-depth-pems07}
\end{table}

\FloatBarrier

\section{Appendix C Embedding Ablation Results}

To examine whether the relative performance of U-Trans and S-Trans depends on the input representation, we conduct embedding ablations across all six datasets. Within each configuration, the two models use exactly the same input embeddings, and only the spatial mixer differs. All results are averaged over five random seeds. The two 24-dimensional settings differ as follows. ``Flow + adaptive only ($d_{\mathrm{adp}}=24$)'' retains only the flow branch and a 24-dimensional adaptive embedding, while removing the congestion-degree and all temporal-embedding branches. ``Full representation ($d_{\mathrm{adp}}=24$)'' retains the complete dataset-specific representation and changes only the adaptive embedding dimension from 80 to 24. Underlined values indicate the better result between U-Trans and S-Trans for each metric within each representation setting.

Interestingly, reducing $d_{\mathrm{adp}}$ from 80 to 24 largely preserves the performance under the full input representation and yields slight improvements on several datasets, suggesting that high-dimensional adaptive embeddings are not necessary for some datasets. Since the dimensional settings largely follow prior models and no dedicated dimensionality search was conducted, we retain a unified 80-dimensional adaptive embedding in the main experiments.

\clearpage
\begin{table}[!t]
\centering
{\small
\renewcommand{\arraystretch}{1.0}
\setlength{\tabcolsep}{1.5pt}
\begin{tabular*}{\textwidth}{@{\extracolsep{\fill}}lcccccc@{}}
\toprule
\multirow{2}{*}{\textbf{Representation setting}}
& \multicolumn{3}{c}{\textbf{U-Trans}}
& \multicolumn{3}{c}{\textbf{S-Trans}} \\
\cmidrule(lr){2-4}
\cmidrule(lr){5-7}
& \textbf{MAE} & \textbf{RMSE} & \textbf{MAPE}
& \textbf{MAE} & \textbf{RMSE} & \textbf{MAPE} \\
\midrule
Flow only & 16.8536 & 28.6987 & 16.2924 & \underline{16.7475} & \underline{28.4943} & \underline{16.1222} \\
w/o TOD embedding & \underline{14.9672} & \underline{26.0696} & \underline{14.4720} & 15.0303 & 26.1551 & 15.1491 \\
w/o adaptive embedding & 15.6744 & 27.1052 & \underline{15.3979} & \underline{15.6642} & \underline{27.0295} & 15.5289 \\
w/o congestion embedding & \underline{14.3615} & \underline{25.2134} & \underline{14.6435} & 14.5451 & 25.7977 & 14.7761 \\
Flow + adaptive only ($d_{\mathrm{adp}}=24$)
& \underline{15.0512} & \underline{26.3792} & \underline{14.6691}
& 15.2813 & 26.8295 & 15.0578 \\
Full representation ($d_{\mathrm{adp}}=24$)
& \underline{14.2553} & \underline{25.1456} & \underline{14.4905}
& 14.3899 & 25.8281 & 14.5583 \\
\bottomrule
\end{tabular*}
}
\caption{Embedding ablation results on PEMS03.}
\label{tab:embedding-pems03}
\end{table}

\begin{table}[!t]
\centering
{\small
\renewcommand{\arraystretch}{1.0}
\setlength{\tabcolsep}{1.5pt}
\begin{tabular*}{\textwidth}{@{\extracolsep{\fill}}lcccccc@{}}
\toprule
\multirow{2}{*}{\textbf{Representation setting}}
& \multicolumn{3}{c}{\textbf{U-Trans}}
& \multicolumn{3}{c}{\textbf{S-Trans}} \\
\cmidrule(lr){2-4}
\cmidrule(lr){5-7}
& \textbf{MAE} & \textbf{RMSE} & \textbf{MAPE}
& \textbf{MAE} & \textbf{RMSE} & \textbf{MAPE} \\
\midrule
Flow only & 22.7667 & 35.9976 & 14.9528 & \underline{22.6401} & \underline{35.8492} & \underline{14.8380} \\
w/o TOD/DOW embeddings & \underline{19.0481} & \underline{30.5963} & \underline{12.5752} & 19.1528 & 30.6974 & 12.6853 \\
w/o adaptive embedding & 21.2454 & 34.2636 & \underline{13.9768} & \underline{21.2320} & \underline{34.2448} & 13.9928 \\
w/o congestion embedding & \underline{18.1316} & \underline{29.9748} & \underline{11.8078} & 18.1438 & 30.0387 & 11.8409 \\
Flow + adaptive only ($d_{\mathrm{adp}}=24$)
& \underline{19.0231} & \underline{30.5729} & \underline{12.6074}
& 19.2890 & 30.9046 & 12.6285 \\
Full representation ($d_{\mathrm{adp}}=24$)
& 18.1613 & \underline{29.9445} & \underline{11.8315}
& \underline{18.1512} & 30.0664 & 11.8733 \\
\bottomrule
\end{tabular*}
}
\caption{Embedding ablation results on PEMS04.}
\label{tab:embedding-pems04}
\end{table}

\begin{table}[!t]
\centering
{\small
\renewcommand{\arraystretch}{1.0}
\setlength{\tabcolsep}{1.5pt}
\begin{tabular*}{\textwidth}{@{\extracolsep{\fill}}lcccccc@{}}
\toprule
\multirow{2}{*}{\textbf{Representation setting}}
& \multicolumn{3}{c}{\textbf{U-Trans}}
& \multicolumn{3}{c}{\textbf{S-Trans}} \\
\cmidrule(lr){2-4}
\cmidrule(lr){5-7}
& \textbf{MAE} & \textbf{RMSE} & \textbf{MAPE}
& \textbf{MAE} & \textbf{RMSE} & \textbf{MAPE} \\
\midrule
Flow only & 23.8395 & 38.1636 & \underline{10.0202} & \underline{23.6787} & \underline{38.0945} & 10.4083 \\
w/o TOD/DOW embeddings & 20.1996 & \underline{32.9968} & 8.5075 & \underline{20.1676} & 33.0105 & \underline{8.4970} \\
w/o adaptive embedding & \underline{21.7118} & \underline{36.1516} & \underline{9.0559} & 21.7824 & 36.2004 & 9.0918 \\
w/o congestion embedding & \underline{18.9484} & \underline{32.4864} & \underline{7.8896} & 19.1793 & 32.7529 & 8.0683 \\
Flow + adaptive only ($d_{\mathrm{adp}}=24$)
& \underline{20.4789} & \underline{33.3647} & 8.5975
& 20.5562 & 33.5410 & \underline{8.5960} \\
Full representation ($d_{\mathrm{adp}}=24$)
& \underline{18.8365} & \underline{32.3218} & \underline{7.8655}
& 18.9221 & 32.3407 & 7.8744 \\
\bottomrule
\end{tabular*}
}
\caption{Embedding ablation results on PEMS07.}
\label{tab:embedding-pems07}
\end{table}

\begin{table}[!t]
\centering
{\small
\renewcommand{\arraystretch}{1.0}
\setlength{\tabcolsep}{1.5pt}
\begin{tabular*}{\textwidth}{@{\extracolsep{\fill}}lcccccc@{}}
\toprule
\multirow{2}{*}{\textbf{Representation setting}}
& \multicolumn{3}{c}{\textbf{U-Trans}}
& \multicolumn{3}{c}{\textbf{S-Trans}} \\
\cmidrule(lr){2-4}
\cmidrule(lr){5-7}
& \textbf{MAE} & \textbf{RMSE} & \textbf{MAPE}
& \textbf{MAE} & \textbf{RMSE} & \textbf{MAPE} \\
\midrule
Flow only & 17.0519 & 27.1740 & \underline{10.7766} & \underline{17.0036} & \underline{27.1489} & 10.7838 \\
w/o TOD/DOW embeddings & \underline{15.0999} & \underline{23.7920} & \underline{9.6975} & 15.3156 & 24.0950 & 9.7386 \\
w/o adaptive embedding & 14.8105 & \underline{25.4279} & \underline{9.6632} & \underline{14.7820} & 25.4572 & 9.6676 \\
w/o congestion embedding & \underline{13.3531} & \underline{23.0717} & \underline{8.8285} & 13.4116 & 23.1464 & 8.8538 \\
Flow + adaptive only ($d_{\mathrm{adp}}=24$)
& \underline{15.2079} & \underline{23.9289} & 9.7849
& 15.4003 & 24.2023 & \underline{9.7174} \\
Full representation ($d_{\mathrm{adp}}=24$)
& \underline{13.3535} & 23.0418 & \underline{8.8087}
& 13.3644 & \underline{22.9880} & 8.8118 \\
\bottomrule
\end{tabular*}
}
\caption{Embedding ablation results on PEMS08.}
\label{tab:embedding-pems08}
\end{table}

\begin{table}[!t]
\centering
{\small
\renewcommand{\arraystretch}{1.0}
\setlength{\tabcolsep}{1.5pt}
\begin{tabular*}{\textwidth}{@{\extracolsep{\fill}}lcccccc@{}}
\toprule
\multirow{2}{*}{\textbf{Representation setting}}
& \multicolumn{3}{c}{\textbf{U-Trans}}
& \multicolumn{3}{c}{\textbf{S-Trans}} \\
\cmidrule(lr){2-4}
\cmidrule(lr){5-7}
& \textbf{MAE} & \textbf{RMSE} & \textbf{MAPE}
& \textbf{MAE} & \textbf{RMSE} & \textbf{MAPE} \\
\midrule
Flow only & 23.8803 & 42.3987 & 18.7011 & \underline{23.3364} & \underline{41.1696} & \underline{18.4967} \\
w/o TOD/DOW embeddings & \underline{17.4607} & \underline{32.1795} & \underline{14.3205} & 17.8069 & 33.4682 & 14.4309 \\
w/o adaptive embedding & \underline{22.3788} & \underline{39.0523} & 18.1646 & 22.4255 & 39.1949 & \underline{18.0699} \\
Flow + adaptive only ($d_{\mathrm{adp}}=24$)
& 17.8060 & \underline{32.6715} & 14.5114
& \underline{17.6944} & 32.7821 & \underline{14.4732} \\
Full representation ($d_{\mathrm{adp}}=24$)
& \underline{16.5481} & \underline{30.2053} & \underline{13.8385}
& 16.7713 & 30.5858 & 13.9749 \\
\bottomrule
\end{tabular*}
}
\caption{Embedding ablation results on T-DRIVE.}
\label{tab:embedding-tdrive}
\end{table}
\clearpage

\begin{table}[!t]
\centering
{\small
\renewcommand{\arraystretch}{1.0}
\setlength{\tabcolsep}{1.5pt}
\begin{tabular*}{\textwidth}{@{\extracolsep{\fill}}lcccccc@{}}
\toprule
\multirow{2}{*}{\textbf{Representation setting}}
& \multicolumn{3}{c}{\textbf{U-Trans}}
& \multicolumn{3}{c}{\textbf{S-Trans}} \\
\cmidrule(lr){2-4}
\cmidrule(lr){5-7}
& \textbf{MAE} & \textbf{RMSE} & \textbf{MAPE}
& \textbf{MAE} & \textbf{RMSE} & \textbf{MAPE} \\
\midrule
Flow only & 14.9205 & 24.3971 & \underline{15.7807} & \underline{14.8261} & \underline{24.2034} & 15.8507 \\
w/o TOD/DOW embeddings & 12.8954 & 21.2912 & 13.5304 & \underline{12.7903} & \underline{21.1408} & \underline{13.4801} \\
w/o adaptive embedding & 13.8611 & 22.5022 & 14.9878 & \underline{13.8578} & \underline{22.4187} & \underline{14.9861} \\
Flow + adaptive only ($d_{\mathrm{adp}}=24$)
& 12.8946 & 21.3328 & 13.5973
& \underline{12.7082} & \underline{21.0162} & \underline{13.4057} \\
Full representation ($d_{\mathrm{adp}}=24$)
& 12.5374 & 20.7662 & 13.2835
& \underline{12.3905} & \underline{20.5269} & \underline{13.1069} \\
\bottomrule
\end{tabular*}
}
\caption{Embedding ablation results on NYCTAXI.}
\label{tab:embedding-nyctaxi}
\end{table}
\FloatBarrier

\section{Appendix D Logit-Level Effect of Front Global-Mean Injection}

Appendix D summarizes the logit-level effect of front global-mean injection. As in the main-paper derivation, projection biases are omitted. The following analysis applies to one attention head. Let $U$ be the row-uniform matrix, and let $S_t=Q_tK_t^{\top}/\sqrt{d_s}$ denote the scaled attention-logit matrix. Compensation changes the query and key projections as follows:
\begin{equation}
  Q'_t=(I+U)Q_t,
  \qquad
  K'_t=(I+U)K_t.
  \tag{D1}
\end{equation}
Starting from the projected queries and keys in Equation (D1), the compensated logits are:
\begin{equation}
  S'_t=(I+U)S_t(I+U)^{\top}
  =S_t+US_t+S_tU+US_tU.
  \tag{D2}
\end{equation}
Although Equation (D2) contains four terms, not all terms change the attention weights after row-wise softmax. Since $U$ is the row-uniform matrix, the third and fourth terms satisfy:
\begin{equation}
  S_tU=\frac{1}{N}S_t\mathbf{1}\mathbf{1}^{\top},
  \qquad
  US_tU=\frac{1}{N^2}\mathbf{1}\mathbf{1}^{\top}S_t\mathbf{1}\mathbf{1}^{\top}.
  \tag{D3}
\end{equation}
Both terms in Equation (D3) are row-wise constant matrices. For each row, they add the same scalar to all columns. Row-wise softmax is invariant to such row-wise constant shifts:
\begin{equation}
  \operatorname{softmax}_{\mathrm{row}}\!\left(X+c\mathbf{1}^{\top}\right)
  =\operatorname{softmax}_{\mathrm{row}}(X).
  \tag{D4}
\end{equation}
Therefore, the effective compensated attention can be written as:
\begin{equation}
  A'_t=\operatorname{softmax}_{\mathrm{row}}(S'_t)
  =\operatorname{softmax}_{\mathrm{row}}(S_t+US_t).
  \tag{D5}
\end{equation}
Consequently, the compensated residual used in Equation (27) is:
\begin{equation}
  \Delta'_t=A'_t-U
  =\operatorname{softmax}_{\mathrm{row}}(S_t+US_t)-U.
  \tag{D6}
\end{equation}
Only the global key-side salience term changes attention; softmax invariance removes the two row-wise constant terms.

\section[Appendix E Complete Results of RI-Trans Interpolation and Front Compensation]{Appendix E Complete Results of RI-Trans Interpolation\\and Front Compensation}

Appendix E reports five-seed mean MAE for the inference-only RI-Trans control, the independently optimized interpolation in Figure 3(a), and the front-compensation profile in Figure 3(b). In the inference-only control, $r$ is varied while all learned parameters remain fixed at the corresponding S-Trans checkpoints. For the independently optimized results, each setting is trained separately under the S-Trans protocol, and relative MAE uses MH-U-Trans as the zero reference, as in Equation (28). All percentage changes are calculated using the unrounded MAE values and rounded to four decimal places for reporting.

\FloatBarrier
\begin{table}[!ht]
\centering
{\small
\setlength{\tabcolsep}{1.5pt}
\begin{tabular*}{\textwidth}{@{\extracolsep{\fill}}ccccccc@{}}
\toprule
\multirow{2}{*}[-0.5ex]{$\mathbf{r}$}
& \multicolumn{2}{c}{\textbf{PEMS03}}
& \multicolumn{2}{c}{\textbf{NYCTAXI}}
& \multicolumn{2}{c}{\textbf{PEMS04}} \\
\cmidrule(lr){2-3}
\cmidrule(lr){4-5}
\cmidrule(lr){6-7}
& \textbf{MAE} & \textbf{Change (\%)}
& \textbf{MAE} & \textbf{Change (\%)}
& \textbf{MAE} & \textbf{Change (\%)} \\
\midrule
0.00 & 20.6821 & $+43.6990$ & 29.7824 & $+139.0641$ & 21.7563 & $+20.0411$ \\
0.25 & 18.0800 & $+25.6195$ & 23.5573 & $+89.0950$ & 20.4021 & $+12.5692$ \\
0.50 & 16.0958 & $+11.8335$ & 18.2049 & $+46.1313$ & 19.2696 & $+6.3205$ \\
0.75 & 14.8466 & $+3.1537$ & 14.1834 & $+13.8504$ & 18.4509 & $+1.8032$ \\
1.00 & 14.3927 & $+0.0000$ & 12.4579 & $+0.0000$ & 18.1241 & $+0.0000$ \\
\bottomrule
\end{tabular*}
}
\caption{Inference-only RI-Trans interpolation results (five-seed mean MAE; percentage changes relative to the $r=1$ endpoint).}
\label{tab:ri-trans-inference-only}
\end{table}

\FloatBarrier
\begin{table}[!ht]
\centering
{\small
\setlength{\tabcolsep}{1.5pt}
\begin{tabular*}{\textwidth}{@{\extracolsep{\fill}}ccccccc@{}}
\toprule
\multirow{2}{*}[-0.5ex]{$\mathbf{r}$}
& \multicolumn{2}{c}{\textbf{PEMS03}}
& \multicolumn{2}{c}{\textbf{NYCTAXI}}
& \multicolumn{2}{c}{\textbf{PEMS04}} \\
\cmidrule(lr){2-3}
\cmidrule(lr){4-5}
\cmidrule(lr){6-7}
& \textbf{MAE} & \textbf{Rel. MAE (\%)}
& \textbf{MAE} & \textbf{Rel. MAE (\%)}
& \textbf{MAE} & \textbf{Rel. MAE (\%)} \\
\midrule
0.00 & 14.2485 & $+0.0000$ & 12.6195 & $+0.0000$ & 18.1499 & $+0.0000$ \\
0.25 & 14.4413 & $+1.3526$ & 12.5036 & $-0.9184$ & 18.1420 & $-0.0433$ \\
0.50 & 14.4927 & $+1.7136$ & 12.5069 & $-0.8923$ & 18.1574 & $+0.0412$ \\
0.75 & 14.4383 & $+1.3317$ & 12.4774 & $-1.1260$ & 18.1178 & $-0.1766$ \\
1.00 & 14.3927 & $+1.0113$ & 12.4579 & $-1.2806$ & 18.1241 & $-0.1424$ \\
\bottomrule
\end{tabular*}
}
\caption{RI-Trans interpolation results.}
\label{tab:ri-trans}
\end{table}

\FloatBarrier
\begin{table}[!ht]
\centering
{\small
\setlength{\tabcolsep}{1.5pt}
\begin{tabular*}{\textwidth}{@{\extracolsep{\fill}}lcccccc@{}}
\toprule
\multirow{2}{*}[-0.5ex]{\textbf{Setting}}
& \multicolumn{2}{c}{\textbf{PEMS03}}
& \multicolumn{2}{c}{\textbf{NYCTAXI}}
& \multicolumn{2}{c}{\textbf{PEMS04}} \\
\cmidrule(lr){2-3}
\cmidrule(lr){4-5}
\cmidrule(lr){6-7}
& \textbf{MAE} & \textbf{Rel. MAE (\%)}
& \textbf{MAE} & \textbf{Rel. MAE (\%)}
& \textbf{MAE} & \textbf{Rel. MAE (\%)} \\
\midrule
MH-U-Trans & 14.2485 & $+0.0000$ & 12.6195 & $+0.0000$ & 18.1499 & $+0.0000$ \\
Front-Comp. & 14.3403 & $+0.6439$ & 12.5423 & $-0.6118$ & 18.1675 & $+0.0968$ \\
S-Trans & 14.3927 & $+1.0113$ & 12.4579 & $-1.2806$ & 18.1241 & $-0.1424$ \\
\bottomrule
\end{tabular*}
}
\caption{Front compensation results.}
\label{tab:front-compensation}
\end{table}
\FloatBarrier

Compared with S-Trans endpoints, front compensation moves relative MAE closer to the MH-U-Trans reference on PEMS03 and NYCTAXI, while PEMS04 stays near zero. The compensated signs remain dataset-dependent, matching the \emph{Residual Diagnostics and Front Global-Mean Injection} analysis in the main paper.

\end{document}